\pdfoutput=1
\documentclass[11pt,a4paper]{article}
\usepackage[hyperref]{eacl2021}
\usepackage{times}
\usepackage{latexsym}
\usepackage{url}
\usepackage{color}
\usepackage{dutchcal}
\usepackage{enumitem}

\usepackage{microtype}

\aclfinalcopy 

\title{Machine Translationese: Effects of Algorithmic Bias on Linguistic Complexity in Machine Translation}

\author{Eva Vanmassenhove$^\alpha$ \\
\And   Dimitar Shterionov$^\alpha$ \\
   $^\alpha$ Cognitive Science and AI, Tilburg University, The Netherlands \\
   { \texttt{\{e.o.j.vanmassenhove, d.shterionov\}}\texttt{@tilburguniversity.edu}} \\
   $^\beta$ University of Maryland, College Park  \\
{ \texttt{mgwillia@umd.edu}} 
\And   Matthew Gwilliam$^\beta$ 
   }

\date{}

\begin{document}
\maketitle
\begin{abstract}
Recent studies in the field of Machine Translation (MT) and Natural Language Processing (NLP) have shown that existing models amplify biases observed in the training data. The amplification of biases in language technology has mainly been examined with respect to specific phenomena, such as gender bias. In this work, we go beyond the study of gender in MT and investigate how bias amplification might affect language in a broader sense. We hypothesize that the `algorithmic bias', i.e. an exacerbation of frequently observed patterns in combination with a loss of less frequent ones, not only exacerbates societal biases present in current datasets but could also lead to an artificially impoverished language: `machine translationese'. We assess the linguistic richness (on a lexical and morphological level) of translations created by different data-driven MT paradigms -- phrase-based statistical (PB-SMT) and neural MT (NMT). Our experiments show that there is a loss of lexical and morphological richness in the translations produced by all investigated MT paradigms for two language pairs (EN$\leftrightarrow$FR and EN$\leftrightarrow$ES).
\end{abstract}

\section{Introduction}~\label{intro}
The idea of translation entailing a transformation is widely recognised in the field of Translation Studies (TS)~\cite{ippolito2014simplification}. Translations are specific communicative acts occurring in a particular context governed by their own laws. Some of the features that characterize translated texts are defined as simplification, explicitation, normalization and leveling out~\cite{baker1999role}. The fingerprints left by the translation process and the language this results into, have been referred to as `translationese'~\cite{gellerstam1986translationese}. Empirical evidence of the existence of translationese can be found in studies showing that machine learning techniques can be employed to automatically distinguish between human translated and original text by looking at lexical and grammatical information~\cite{baroni2006new,koppel2011translationese}. Translationese differs from original texts due to a combination of factors including intentional (e.g. explicitation and normalization) and unintentional ones (e.g. unconscious effects of the source language input on the target language produced). 
%From here on, we will refer to this type of Translationese as `Human Translationese'.
Unlike other work on (human) translationese (or even related work on `Post-editese'), we delve into the effects of machine translation (MT) algorithms on language, i.e. `machine translationese'. 

So far, generating accurate and fluent translations has been the main objective of MT systems. As such, maintaining the richness and diversity in the outputs has understandably not been a priority~\cite{vanmassenhove2020integration}.\footnote{One might argue that for some tasks and domains, diversity can be harmful (specific in-domain translations would prefer consistency over e.g. lexical diversity).} However, as MT systems have reached a quality that is (arguably) close to that of human translations~\cite{laubli2018has,toral2018attaining} and as such are being used widely on a daily basis, we believe it is time to look into the potential effects of (MT) algorithms on language itself.\footnote{Google Translate alone translates more than 100 billions words per day and is used by at least 500 million people according to estimates (\url{https://www.blog.google/products/translate/ten-years-of-google-translate/}).}

The main motivations behind this work are: (i) if algorithmic bias is indeed a by-product of our algorithms, a statistically biased MT system might prefer frequently occurring words (or sub-words) over others. Since MT systems do not necessarily distinguish between different synonymous translations (lexical richness) and morphological variants (grammatical richness), algorithmic bias could lead to the loss of morphological variety (and thus interfere with the ability of our systems to generate at all times a grammatically correct option); (ii) the sociolinguistic perspective of machine translationese since it has been established that language contact (e.g. via translationese) can entail language changes~\cite{kranich2014translations}. If machine translationese (and other types of `NLPese') is a simplified version of the training data, what does that imply from a sociolinguistic perspective and how could this affect language on a longer term?

The main objective of the presented research is to establish whether there is indeed a quantitatively measurable difference between the linguistic richness of an MT system's training data and its output in terms of morphological and lexical diversity. To do so, we conduct an in-depth analysis that goes beyond frequently used standard lexical diversity metrics such as TTR, Yule's I and MTLD. We assess the lexical and morphological diversity through an adapted version of the Lexical Frequency Profile used to assess language acquisition, a measure of morphological diversity based on Shannon and Simpson Diversity and an novel automatic evaluation of synonym frequency. %While we focus on state-of-the-art MT systems, we also experiment with Phrase-Based Statistical MT (PB-SMT) for completeness since this is, up to our knowledge, the first study of this nature.
We focus on the most prominent data-driven MT paradigms: Neural MT (NMT), both LSTM~\cite{bahdanau2014neural,Sutskever2014} and Transformer~\cite{Vaswani2017}, and Phrase-Based Statistical MT (PB-SMT). Up to our knowledge this is the first research on lexical and morphological diversity of machine translation output, i.e. machine translationese.

The contributions of this work can be summarised as: (i) a detailed analysis of lexical and morphological diversity of machine translationese and the loss thereof to quantify the effects of algorithmic bias; (ii) the adaptation of a metric used in language acquisition for assessing lexical sophistication in MT\footnote{In fact, our implementation of the LFP metric can be employed for any NLP tasks.}; (iii) the use of Shannon entropy and Simpson diversity to measure morphological richness, and (iv) a novel, automatic evaluation of synonym frequency.

\section{Related Work}\label{sec:relwork}

Several studies have exposed the societal biases present in datasets (racial bias~\cite{merullo2019investigating}, political bias~\cite{fan2019plain}, gender bias~\cite{10045_76104}. Existing NLP technology are likely to pick up biases present in the training data and various explorations of e.g. gender bias in NLP systems have indeed revealed the existence of harmful biases in the output they generate \cite{bolukbasi2016man,caliskan2016semantics,garg2018word,vanmassenhove2019getting,stanovsky-etal-2019-evaluating,sun2019mitigating,habash2019automatic}. Research related to bias has often focused on gender or race. Especially in a field such as MT, the implicit gender in a language such as English and its consecutive translations into morphologically richer languages with gender agreement, makes it relatively easy to expose and study biases related to gender in a contrastive linguistic setting. In the context of this paper, we would like to note that (statistical) bias is not limited to gender or race but can be defined as any systematic inaccuracy in one direction leading to an under (or over) estimation of observations. 

A handful of recent work has mentioned the possibility of algorithmic bias on top of the already existing societal biases in the training data~\cite{bolukbasi2016man,caliskan2016semantics,garg2018word}. For instance, Zhao et al.~\shortcite{zhao-etal-2017-learning} observe a phenomenon they refer to as `bias amplification'. They note that in their training data an activity such as `cooking' is associated 33\% times more with women compared to men. After training a model on that dataset, the existing disparity is amplified to 68\% times more associations with women. 

In the field of MT, Vanmassenhove et al.~\shortcite{vanmassenhove-etal-2019-lost} address the effects of statistical bias on language generation in an MT setting. They assess lexical diversity using standard metrics --TTR, MTLD and Yule's K-- and conclude that the translations produced by various MT systems (PB-SMT and NMT) are consistently less diverse than the original training data. Their approach was conducted on NMT systems that were trained without byte-pair-encoding (BPE)~\cite{sennrich-etal-2016-bpe} which limits the creativity of the translation systems.

Toral~\shortcite{toral2019post} measures the lexical diversity of 18 state-of-the-art systems on 6 language pairs, reaching similar conclusions. They do so focusing specifically on post-editese. The experiments indicate that post-editese is simpler  and  more normalised than human translationese. The post-editese also shows a higher degree of interference from the source compared to the human translations. Daems et al.~\shortcite{daems2017translationese}, like Toral~\shortcite{toral2019post}, centers around the automatic detection of post-editese and does not look into properties of unedited machine translationese. In Aranberri ~\cite{aranberri2020can} different freely available MT systems (neural and rule-based) are compared in terms of automatic metrics (BLEU, TER) and translationese features (TTR, length ratio input{/}output, perplexity, etc.) to investigate how such features correlate with translation quality. Bizzoni et al. \shortcite{van2020human} presents a comparison using similar translationese features of three MT architectures and the human translations of spoken and written language.

In the field of PB-SMT, Klebanov and Flor~\shortcite{klebanov2013associative} show that PB-SMT suffers considerably more than human translations (HT) from lexical loss, resulting in loss of lexical tightness and text cohesion. Aharoni et al.~\shortcite{aharoni2014automatic} proof that automatic and human translated sentences can be automatically identified corroborating that human translations systematically differ from the translations produced by PB-SMT systems.

Aside from Vanmassenhove et al.~\shortcite{vanmassenhove-etal-2019-lost}, the above discussed related work uses metrics of lexical diversity to compare human translations to (post-edited) machine translations. In this work, we compare how and whether the output of an MT system differs (in terms of lexical and morphological diversity) from the data it was originally trained on. This way, we aim to investigate the effect of the algorithm (and algorithmic bias) on language itself.

\section{Machine Translation Systems}\label{sec:mt}
MT paradigms have changed quickly over the last decades. Since this is the first attempt to quantify both the lexical and grammatical diversity of machine translationese, we experimented with the current state-of-the-art data-driven paradigms, LSTM and Transformer, as well as with PB-SMT. We used data from the Europarl corpus~\cite{Koehn2005} for two language pairs, English--French and English--Spanish in both direction (EN$\rightarrow$FR, FR$\rightarrow$EN, EN$\rightarrow$ES and ES$\rightarrow$EN). %This was done in order to verify whether there would be a difference in terms of (the potential loss of) diversity when comparing morphologically richer languages (French and Spanish) to a morphologically poorer one (English).
We are interested in both directions in order to verify whether there is a difference in terms of (the potential loss of) diversity when comparing translations from a morphologically poorer language (English) into morphologically richer ones (French and Spanish) and vice versa. Our data is summarised in Table~\ref{tbl:data_stats}.\footnote{We ought to address the fact that the Europarl data consists of both human-uttered and translated text which have different properties in terms of diversity. In this work we analyse the impoverishment of data when it passes through the ``filter'' of the MT system, i.e. the effect of algorithm. As the origin of the data, human-uttered or translated, has no impact on the inherent workings of the MT system we do not take this into account in our analysis.} 

\begin{table}[htb]
\centering
{\small
    \begin{tabular}{|c|c|c|c|}\hline
        Lang. pair & Train & Test & Dev \\\hline
        EN-FR/FR-EN & 1,467,489 & 499,487 & 7,723\\%\hline
        EN-ES/ES-EN & 1,472,203 & 459,633 & 5,734\\\hline
    \end{tabular}}
    \caption{Number of parallel sentences for the training, testing and development sets.}
    \label{tbl:data_stats}
\end{table}

The specifics of the MT systems we trained are:

\noindent\textbf{PB-SMT~~} For the PB-SMT systems we used Moses~\cite{Koehn2007} with default settings and a 5-gram language model with pruning of bigrams. We also tuned each system using MERT~\cite{Och2003} until convergence or for a maximum of 25 iterations. During translation we mask unknown words with the UNK token to avoid bleeding through (source) words which would artificially increase the linguistic diversity. %During translation we mark unknown words to allow us to identify which words are bleeding through.

\noindent\textbf{NMT~~} For the RNN and Transformer systems we used OpenNMT-py.\footnote{\url{https://opennmt.net/OpenNMT-py/}} The systems were trained for maximum of 150K steps, saving an intermediate model every 5000 steps or until reaching convergence according to an early stopping criteria of no improvements of the perplexity (scored on the development set) for 5 intermediate models. The options we used for the neural systems are: 
\setlist{nolistsep, leftmargin=*}
\begin{itemize}[noitemsep]
    \item RNN: size: 512, RNN type: bidirectional LSTM, number of layers of the encoder and of the decoder: 4, attention type: MLP, dropout: 0.2, batch size: 128, learning optimizer: Adam~\cite{Kingma2014} and learning rate: 0.0001.
    \item Transformer: number of layers: 6, size: 512, transformer\_ff: 2048, number of heads: 8, dropout: 0.1, batch size: 4096, batch type: tokens, learning optimizer Adam with beta$_2 = 0.998$, learning rate: 2.
\end{itemize}
All NMT systems have the learning rate decay enabled and their training is distributed over 4 nVidia 1080Ti GPUs. The selected settings for the RNN systems are optimal according to Britz et al.~\shortcite{Britz2017}; for the Transformer we use the settings suggested by the OpenNMT community\footnote{\url{http://opennmt.net/OpenNMT-py/FAQ.html}} as the optimal ones that lead to quality on par with the original Transformer work~\cite{Vaswani2017}. 

For training, testing and validation of the systems we used the same data. To build the vocabularies for the NMT systems we used sub-word units, allowing NMT to be more creative; using sub-word units also mitigates to a certain extent the out of vocabulary problem. To compute the sub-word units we used BPE with 50,000 merging operations for all our data sets. 
% We ought to stress that for both RNN and Transformer we used exactly the same data; for PB-SMT the data was not split in sub-word units to meet the conditions under which optimal results have been achieved for both paradigms.
In Table~\ref{tbl:data_stats2} we present the vocabulary sizes of the data used to train our PB-SMT and NMT systems.
\begin{table}[htb]
\centering
{\small
    \begin{tabular}{|c|c|c|c|c|}\hline
         & \multicolumn{2}{c|}{no BPE} & \multicolumn{2}{c|}{with BPE} \\\cline{2-5}
        Lang. pair & EN & FR/ES & EN & FR/ES \\\hline
        EN-FR/FR-EN & 113,132 & 131,104 & 47,628 & 48,459\\
        EN-ES/ES-EN & 113,692 & 168,195 & 47,639 & 49,283\\\hline
    \end{tabular}}
    \caption{Vocabulary sizes. For completeness we also present the vocabulary size without BPE, i.e. the number of unique words in the corpora.}
    \label{tbl:data_stats2}
\end{table}

The quality of our MT systems is evaluated on the test set using standard evaluation metrics -- BLEU~\cite{papineni2002BLEU} (as implemented in SacreBLEU~\cite{post-2018-call}) and TER~\cite{Snover2006} (as implemented in Multeval~\cite{Clark2011}). Our evaluation scores are presented in Table~\ref{tbl:mt_eval}. 

\begin{table}[thb]
\centering
{\small
    \begin{tabular}{|l|c|c|c|c|}
    \multicolumn{5}{c}{English as source}\\\hline
    & \multicolumn{2}{c|}{EN$\rightarrow$FR} & \multicolumn{2}{c|}{EN$\rightarrow$ES} \\\cline{2-5}
    System  & BLEU$\uparrow$ & TER$\downarrow$ & BLEU$\uparrow$ & TER$\downarrow$ \\\hline
PB-SMT & 35.7 & 50.7 & 38.6 & 45.9 \\\hline
LSTM & 34.2 & 50.9 & 38.2 & 45.3 \\\hline
TRANS & \textbf{37.2} & \textbf{48.7} & \textbf{40.9} & \textbf{43.4} \\\hline
\multicolumn{5}{c}{English as target}\\\hline
System & \multicolumn{2}{c|}{FR$\rightarrow$EN} & \multicolumn{2}{c|}{ES$\rightarrow$EN}\\ \hline
PB-SMT & 36.2 & 47.1 & 39.3 & 44.0\\\hline
LSTM & 34.6 & 48.2 & 38.1 & 44.7\\\hline
TRANS & \textbf{37.0} & \textbf{46.4} & \textbf{41.3} & \textbf{41.4}\\\hline
    \end{tabular}
    }
    \caption{Quality evaluation scores for our MT systems. TRANS denotes Transformer systems.}
    \label{tbl:mt_eval}
\end{table}
We computed pairwise statistical significance using bootstrap resampling~\cite{koehn-2004-statistical} and a 95\% confidence interval. The results shown in Table~\ref{tbl:mt_eval} are all statistically significant based on 1000 iterations and samples of 100 sentences. All metrics show the same performance trends for all language pairs: Transformer (TRANS) outperforms all other systems, followed by PB-SMT, and LSTM.

For all PB-SMT we replaced marked unknown words with only one token ``UNK''. While this does not effect the computation of BLEU and TER, it allows us not to artificially boost the lexical and grammatical scores for these MT engines (see Section~\ref{sec:exp}) and assess their realistic dimensions.

\section{Experiments and Results}~\label{sec:exp}
Assessing linguistic complexity is a multifaceted task spanning over various domains (lexis, morphology, syntax, etc.). The lexical and grammatical diversity are two of its major components~\cite{bachman2004statistical,bulte2008investigating,bulte2013development}. As such, we conduct an analysis using (i) lexical diversity and sophistication metrics (Section~\ref{subsec:lexdiv}) and (ii) grammatical diversity metrics (Section~\ref{subsec:gramdiv}). For lexical diversity, we use the following metrics: an adapted version of the Lexical Frequency Profile (LFP), three standard metrics commonly used to assess diversity --TTR, Yule's I and MTLD--, and three new metrics based on synonym frequency in translations. Up to our knowledge, this research is the first to employ LFP to analyze synthetic data. For grammatical diversity, we focus specifically on morphological inflectional diversity. We adopt the Shannon entropy and Simpson's diversity index to compute the entropy of the inflectional paradigms of lemmas, measuring the abundance and the evenness of wordforms per lemma.

Next, we will discuss the evaluation data and the aforementioned metrics designed in order to compare the diversity of the training data with the machine translationese. Our evaluation scripts are available at \url{https://github.com/dimitarsh1/BiasMT}; due to its large size, the data is not hosted in the github repository but is available upon request.

\subsection{Evaluation data}
To observe the effects of the MT algorithm on linguistic diversity, we used the MT engines (Section~\ref{sec:mt}) to translate the source side of the training set, i.e. completely observed data. Data that has fully been observed during training is most suitable for our objectives as we are interested in the effects of the algorithm on language itself. 
It is also the most favourable translation (and evaluation) scenario for the MT systems since all data has been observed.

\subsection{Lexical Diversity}\label{subsec:lexdiv}
\paragraph{Lexical Frequency Profile}
To look at the lexical sophistication and diversity in the text produced by the MT systems, we adapted the Lexical Frequency Profile (LFP) method~\cite{laufer1994lexical,laufer1995vocabulary}. LFP is a measure that stems from research in second language (L2) acquisition and student writing methods. It is designed to study the lexical diversity or sophistication in texts produced by L2 learners. It is based on the observation that texts including a higher proportion of less frequent words are more sophisticated than those containing higher proportions of more frequent words~\cite{kyle2019measuring}.
    
The LFP method measures diversity and sophistication by looking at frequency bands. In its original version, the LFP analysis would distinguish between 4 bands: (i) percentage of words in a text belonging to the 1000 most frequent words in that language, (ii) percentage of words in a text belonging to the next 1000 most frequent words, (iii) a list of academic words that did not occur in the first 2000 words and (iv) the remaining words. The lists used to determine the word bands are predefined word lists such as Nation's word lists~\cite{nation1984vocabulary}. One shortcoming of the approach is that a mismatch between reference corpus and the target text can lead to misleading outcomes. However, since we are looking into the (side-)effects of the training algorithm, instead of using preset word lists in order to compute the LFP, we use the original training data to generate the word frequency lists. This allows for a better comparison between the original data and the machine translationese while bypassing the potential mismatch issue.

Several studies~\cite{crossley2013comparing,laufer1994lexical,laufer1995vocabulary} have employed the LFP method to assess L2 acquisition in learners. From these studies, it resulted that the less proficient a user of an L2, the more words belonged to the first band and the least words belong to the list of academic words or the remaining words (band 3 and 4 respectively of the original formulation). 
    
The lexical profile mentioned above is a detailed profile, showing 4 types of words used by the learner. Because of interpretability issues, the `Beyond 2000' is also frequently used to assess the profile of the users. It distinguishes between the first two bands (comprising of the first 2000 words) and the rest. This condensed profile has been found equally reliable and valid as the original LFP having the advantage that it reduces the profile to one single score facilitating a comparison between learners (or in our case MT systems).
    
Since we are interested in the difference between the original training data and the output of the MT systems, we compute the frequency bands on the original training data instead of based on pre-set word lists used in L2 research. As such, we leave out the third band consisting of a list of academic words. Presenting and computing the LFP this way, will give us immediately the `Beyond 2000' metric score as well (as we distinguish between three bands only, the last one being anything beyond the first 2000 words).

The LFP for French, Spanish and English, from the EN$\rightarrow$FR, EN$\rightarrow$ES, FR$\rightarrow$EN (denoted as EN$_{FR}$) and ES$\rightarrow$EN (denoted as EN$_{ES}$) data is presented in Table~\ref{tbl:LFPfrenchspanishenglish}. 

It shows that the original data is consistently more diverse than the output of the MT systems as (i) the percentage of text occupied by the 1000 most frequent words (B1) is lower than in the corresponding B1 scores for all MT systems which implies that the 1000 most frequent words take up a smaller percentage of the text in the original training data compared to in the output of the different MT systems; and (ii) the so-called `Beyond 2000' measure, which in our LFP is equal to the third band (B3), showing us the percentage of text occupied by the words that do not belong to the first two bands, is consistently higher for the original data compared to the MT systems (meaning that the less frequent words occupy a bigger proportion of the original data than they do in its machine translationese variants). Note that it has been established that LFPs are large-grained so small gains in vocabulary are likely to be obscured~\cite{kyle2019measuring}. The results indicate a consistent and clear difference between the original data and the different types of machine translationese for all language pairs. 

Aside from the different LFP scores between the training data and the translations, we also see a difference between the languages themselves. French and Spanish have more variety (higher B1 and lower B3 (Beyond 2000) values) compared to EN$_{ES}$ and EN$_{FR}$. Since the LFPs are computed on tokens, this reflects the richer morphology in French and Spanish compared to English.

\begin{table}
\centering
{\small \setlength\tabcolsep{5pt} \begin{tabular}{|l|r|r|r|r|r|r|} \hline
& \multicolumn{3}{c|}{FR} & \multicolumn{3}{c|}{ES} \\\cline{2-7} 
& B1 & B2 & B3 & B1  & B2  & B3 \\\hline
ORIG & \bf{79.80} & 6.59  & \underline{13.61}   & \bf{77.80} & 6.83  & \underline{15.36} \\\hline
PB-SMT  & 81.78 & 6.48  & 11.74 & 79.77 & 6.86  & 13.36  \\\hline
LSTM & 82.95 & 6.18  & 10.88 & 80.34 & 6.84  & 12.81 \\\hline
TRANS & 82.01 & 6.24  & 11.75 & 82.35 & 6.99  & 10.67 \\\hline\hline

& \multicolumn{3}{c|}{EN$_{FR}$} & \multicolumn{3}{c|}{EN$_{ES}$}\\\cline{2-7}
& B1  & B2  & B3 & B1  & B2  & B3 \\\hline
ORIG & \textbf{80.83} & 7.10  & \underline{12.07} & \bf{80.81} & 7.11  & \underline{12.08}\\\hline
PB-SMT & 82.06 & 7.04  & 10.90 & 82.25 & 7.01  & 10.74\\\hline
LSTM & 83.23 & 6.93  & 9.81 & 83.29 & 6.93  & 9.78 \\\hline
TRANS & 82.25 & 7.05  & 10.70 & 82.35 & 6.99  & 10.67\\\hline
\end{tabular}
    \caption{Lexical Frequency Profile (French, Spanish, English (EN$_{FR}$ and EN$_{ES}$) with 3 bands (B1: 0-1000, B2: 1001-2000, B3: 2001-end) for the original data and the output of the MT systems.}
    \label{tbl:LFPfrenchspanishenglish}}
\end{table}

\paragraph{TTR, Yule's I and MTLD} For completeness, we also present three more commonly used measures of lexical diversity: type/token ratio (TTR)~\cite{Templin1975certain}, Yule's K (in practice, we use the reverse Yule's I)~\cite{Yule1944}, and the measure of textual lexical diversity (MTLD)~\cite{Mccarthy2005assessment}. 

TTR presents the ratio of the total number of \textit{different} words (types) to the \textit{total} number of words (tokens). Higher TTR indicates a higher degree of lexical diversity. 

Yule's characteristic constant (Yule's K)~\cite{Yule1944} measures constancy of text as the repetitiveness of vocabulary. Yule's K and its inverse Yule's I are considered to be more resilient to fluctuations related to text length than TTR~\cite{oakes2012}. The third lexical diversity metric is MTLD. MTLD is evaluated sequentially as the mean length of sequential word strings in a text that maintains a given TTR value~\cite{Mccarthy2005assessment}.\footnote{In our experiments we used 0.72 as a TTR threshold.}
We present the scores for TTR, Yule's I and MTLD for our data and MT engines in Table~\ref{tbl:lex_richness}.

\begin{table}[h]
\centering
{\small \setlength\tabcolsep{3.3pt} \begin{tabular}{|l|r|r|r|r|r|r|} \hline
& \multicolumn{3}{c|}{FR} & \multicolumn{3}{c|}{ES} \\\cline{2-7} 
 & TTR & Yule's I & MTLD & TTR & Yule's I & MTLD \\\hline
ORIG & \textbf{3.02} & \textbf{9.28} & \textbf{119.40} & \textbf{4.08} & \textbf{12.31} & \textbf{96.23} \\\hline
PB-SMT & 1.79 & 3.00 & 112.00 & 2.37 & 4.02 & 92.01 \\\hline
LSTM & 1.56 & 2.14 & 104.89 & 2.03 & 2.95 & 86.57 \\\hline
TRANS & 2.07 & 3.82 & 115.66 & 2.89 & 6.23 & 95.72 \\\hline\hline

& \multicolumn{3}{c|}{EN$_{FR}$} & \multicolumn{3}{c|}{EN$_{ES}$}\\\cline{2-7}
& TTR & Yule's I & MTLD & TTR & Yule's I & MTLD\\\hline
ORIG & \textbf{2.89} & \textbf{6.64} & \textbf{108.70} & \textbf{2.88} & \textbf{6.61} & \textbf{108.46} \\\hline
PB-SMT & 1.74 & 2.07 & 94.65 & 1.82 & 2.25 & 93.18\\\hline
LSTM & 1.50 & 1.53 & 86.93 & 1.44 & 1.42 & 87.91 \\\hline
TRANS & 2.04 & 3.10 & 101.95 & 2.09 & 3.26 & 99.62\\\hline
\end{tabular}
    \caption{TTR, Yule's I and MTLD scores. For all metrics, higher scores indicate higher lexical richness. For ease of readability and comparison we multiplied TTR scores by 1,000 and Yule's I scores by 10,000.}
    \label{tbl:lex_richness}}
\end{table}

The scores in Table~\ref{tbl:lex_richness} show that, overall, and according to all three metrics, the original training data has a higher lexical diversity than the machine translationese. 

The data for the morphologically richer languages (FR, ES) as well as its machine translationese variants (PB-SMT, LSTM and TRANS) have higher lexical richness than the morphologically poor(er) language (EN).

\paragraph{Synonym Frequency Analysis}
\begin{table}
\centering
{\small  \setlength\tabcolsep{2pt} 
    \begin{tabular}{|l|c|c|c|c|c|c|}\hline
& \multicolumn{3}{c|}{FR} & \multicolumn{3}{c|}{ES} \\\cline{2-7}
& PTF$\downarrow$ & CDU$\downarrow$ & SynTTR$\uparrow$ & PTF$\downarrow$ & CDU$\downarrow$ & SynTTR$\uparrow$ \\\hline
ORIG & \textbf{9.666} & \textbf{2.725} & \textbf{15.10} & \textbf{9.131} & \textbf{4.539} & \textbf{21.13}\\\hline
PB-SMT & 9.715 & 2.957 & 11.87 & 9.236 & 4.637 & 17.4\\\hline
LSTM & 9.748 & 3.154 & 10.96 & 9.32 & 4.782 & 15.34\\\hline
TRANS & 9.717 & 3.077 & 12.25 & 9.285 & 4.687 & 17.15\\\hline

    \end{tabular}
    }
    \caption{Synonym frequency metrics for our MT systems: primary translation frequency (PTF), cosine distance from uniform (CDU) and TTR modified to only consider words with multiple translation options (SynTTR). The SynTTR scores were multiplied by 100,00 for easier viewing. Higher SynTTR scores indicate greater diversity, while lower PTF and CDU scores indicate greater diversity.}
    \label{tbl:syn_freq}
\end{table}

The objective of synonym frequency analysis is to understand, for words with multiple possible translations, with what frequency the various translations for a given word appear in the translated text. It is called \textit{synonym} frequency in reference to the fact that when translating from one language to another, it is common for a word in the source language to have a corresponding word in the target language to which the source word is typically translated, and that primary translated word can have many \textit{synonyms} that constitute acceptable alternative translation options. Note that we perform this analysis only in one direction: from English into the morphologically richer languages French and Spanish.

To examine synonym frequency, we first lemmatize the text using SpaCy.\footnote{https://spacy.io/} Next, we map all nouns, verbs, and adjectives in the source to their possible translation options retrieved from bilingual dictionaries.\footnote{English-Spanish: https://github.com/mananoreboton/en-es-en-Dic, English-French: https://freedict.org/downloads}

We then count the number of appearances of these different translation options for the ORIG as well as the MT data. 
For example, for the English word ``look'' with translation options in Spanish \{``mirar'', ``esperar'', ``buscar'', ``parecer'', ``dar'', ``vistazo'', ``aspecto'', ``ojeada'', ``mirada''\}, the number of appearances in the TRANS data are as follows: \{(``mirar'': 4002), (``esperar'': 3302), (``buscar'': 2814), (``parecer'': 1144), (``dar'': 977), (``vistazo'': 182), (``aspecto'': 46), (``ojeada'': 0), (``mirada'': 0)\}.
From this mapping of translation option to number of appearances we take a vector consisting only of the numbers of appearances for each translation option, and refer to this as a translated word distribution. That is, for the aforementioned example, the distribution vector is: \{4002, 3302, 2014, 1144, 997, 182, 46, 0, 0\}.
We use these counts and distributions as described below. 

Our first synonym frequency metric deals directly with the primary translation frequency (PTF), where the ``primary translation'' is the translation for a given source word that appears in the target text most often. We argue that selecting secondary translation options for each source word less frequently, and selecting the primary option more frequently, indicates a decrease of lexical diversity. We measure the PTF by taking the average \textit{primary translation prevalence} over all source words for each MT system.

As a second metric we used the cosine distance between a uniform translated word distribution, where each translation option would be equally prevalent, and the actual translation distributions (we denote this metric as CDU). While the ideal distribution of translations for a given word is almost certainly non-uniform (and therefore not perfectly diverse), this metric still gives valuable information about the tendencies of different systems to favor certain translation options over others.

The third metric is a modified TTR which we refer to as Synonym TTR (or SynTTR). Unlike with regular type/token ratio, rather than considering all tokens that appear in the text, we consider as types only translation options from the source-target mappings described above and as tokens we consider only appearances of valid types. This metric exposes where translation systems completely drop viable translation options from their vocabulary.

Table~\ref{tbl:syn_freq} shows the results for these 3 metrics. Interestingly, the MT systems can be ranked in the same order according to all these metrics: PB-SMT $>$ TRANS $>$ LSTM, where $>$ denotes the comparison of lexical diversity (higher to lower). However, across the 3 metrics, for both language pairs, the reference translations (ORIG) appear to be the most lexically diverse in terms of synonym frequency, with the lowest PTF and CDU and highest SynTTR. This reinforces the idea that MT algorithms have a negative impact on the diversity of language.

\subsection{Grammatical Diversity}\label{subsec:gramdiv}
Grammatical diversity manifests itself on the sentence (syntactic complexity) and word level (morphological complexity). With our experiments, we focus on the morphological complexity by averaging the inflectional diversity of all lemmas. To do so, we adopted two measures, originating from Information Theory: Shannon entropy~\cite{shannon1948mathematical} and Simpson's Diversity Index~\cite{simpson1949measurement}. The former emphasizes on the richness aspect of diversity while the latter on the evenness aspect of diversity. We used the Spacy-udpipe lemmatizer to retrieve all lemmas.\footnote{\url{https://github.com/TakeLab/spacy-udpipe}}

\paragraph{Shannon Entropy}
Shannon entropy ($H$) measures the level of uncertainty associated with a random variable ($X$). 
It has been applied in use-cases from economy, ecology, biology, complex systems, language and many others~\cite{Page2007difference,Page2011Diversity_and_Complexity}. In the study of language Shannon entropy has previously been used for estimating the entropy of language models~\cite{Behr2003comparingentropy}. 
We  use it to measure the entropy of wordforms given a lemma. In particular, the entropy of inflectional paradigm of a specific lemma could be computed by taking the base frequency of that lemma (frequency of all wordforms associated with that lemma) and the probabilities of all the wordforms within the inflectional paradigm of that particular lemma. Using such a formulation of entropy allows us to measure the morphological variety (or the loss thereof) for the machine translationese produced by each system -- higher values of $H$ indicate higher diversity and vice-versa.

We use Equation~\ref{eq:shannon_lemma} to compute the entropy of the inflectional paradigm of a lemma. 

\begin{equation}\label{eq:shannon_lemma}
H(\mathcal{l}) = -\sum\limits_{\mathcal{wf} \in \mathcal{l}} p(\mathcal{wf} | \mathcal{l})\log p(\mathcal{wf | \mathcal{l})})   
\end{equation}
\noindent $H(\mathcal{l})$ denotes the entropy of the lemma $\mathcal{l}$ and, for the wordform $\mathcal{wf}$, $p(\mathcal{wf}|\mathcal{l})$ is computed as the fraction of the counts of the wordform, $count(\mathcal{wf})$, to the count of all wordforms for the lemma $\mathcal{l}$, i.e. $p(\mathcal{wf}|\mathcal{l}) = \frac{count(\mathcal{wf})}{\sum_{wf^{*} \in \mathcal{l}}count(wf^{*})}$. We use $\in$ to indicate wordforms of a given lemma.

\paragraph{Simpson's Diversity Index}
Like Shannon Entropy, Simpson's Diversity Index ($D$) is a measure used to determine variation in categorical data. Values close to 1 indicate higher homogeneity, thus lower diversity and values close to 0 indicate higher variability, thus higher diversity.  

Following the same reasoning as with Shannon entropy, we compute Simpson's diversity index for each lemma and the corresponding wordforms according to the formula in Equation~\ref{eq:simpson}.
\begin{equation}\label{eq:simpson}
    D(\mathcal{l}) = \frac{1}{\sum\limits_{\mathcal{wf} \in \mathcal{l}}{p(\mathcal{wf} | \mathcal{l})^2}}
\end{equation}
We average the Shannon entropy and Simpson's diversity index for all lemmas to get an indicative score for each translation system or the original text. We denote these with \textbf{\em{H}} and \textbf{\em{D}}, accordingly. To the best of our knowledge, our work is the first to use Shannon entropy and Simpson's diversity index for the study of inflectional richness on a text level. The closest to our application of these two diversity metrics for measuring inflectional richness is the work by del Prado Mart\'{i}n et al.~\shortcite{del2004putting}. Their work on morphological processing uses Shannon entropy to compute the amount of information carried by a morphological paradigm.

An illustration of the Shannon entropy and Simpson's diversity index of a lemma is given in Table~\ref{tbl:ex_richness}. We list the number of occurrences for every wordform (male singular, male plural, female singular and female plural) appearing in our datasets for the French lemma `pr\'{e}sident' (EN: president). We then compute $H$ and $D$ by applying Equation~\ref{eq:shannon_lemma} and Equation~\ref{eq:simpson} accordingly. %resulting for the original dataset in a score of 18.11 and 92.95 respectively.
While both Shannon $H$ and Simpson's $D$ scores usually range between 0--1, for ease of readability we multiply the scores presented in Table~\ref{tbl:ex_richness} and Table~\ref{tbl:Shannonfrenchspanishenglish} by 100 to present them in the range of $[0-100]$.

\begin{table}[h]
\centering
{\small \setlength\tabcolsep{2.3pt} \begin{tabular}{|l|r|r|r|r|c|c|}\hline
& \multicolumn{4}{l|}{lemma: pr\'{e}sident} & & \\\cline{2-5}
  &  {\scriptsize pr\'{e}sident}  &  {\scriptsize pr\'{e}sidents}  &  {\scriptsize pr\'{e}sidente}  &  {\scriptsize pr\'{e}sidentes}  &  H$\uparrow$  &  D$\downarrow$ \\\hline
ORIG  & 93774 & 2029 & 1490 & 8 &  \textbf{18.11}  &  \textbf{92.95} \\\hline
PB-SMT  & 99367 & 2019 & 496 & 1 & 12.81 & 95.16\\\hline
LSTM  & 95272 & 2039 & 291 &  N/A  & 12.17 & 95.3\\\hline
TRANS  & 92946 & 1952 & 617 &  N/A  & 13.86 & 94.74\\\hline
\end{tabular}
    \caption{An illustration of the Shannon entropy and the Simpson's diversity index computed for the occurrences of the different wordforms of the French lemma for `president' (pr\'{e}sident).}
    \label{tbl:ex_richness}}
\end{table}

For lemmas with a single wordform Shannon entropy and Simpson's diversity index will be $H = 0.0$ and $D = 1.0$, respectively. While this makes sense when measuring the diversity of one morphological paradigm, they actually impact the average scores \textbf{\em{H}} and \textbf{\em{D}} without contributing to the understanding of diversity in a comparative study such as ours. In particular, lemmas with single wordforms may be either an evidence of low diversity, e.g. a translation system will always generate only one form or of high diversity, e.g. rare words that are single wordform for a particular lemma (such as synonyms of more common words) can indicate the ability of a system to generate more diverse language (in terms of synonymy). That is why we computed \textbf{\em{H}} and \textbf{\em{D}} on lemmas with two or more wordforms. For completeness, we also present the number of single wordform lemmas.
The Shannon entropy and Simpson's diversity index for French, Spanish and English for all datasets are presented in Table~\ref{tbl:Shannonfrenchspanishenglish}. The scores are shown in the range $[0 - 100]$ as noted above.

\begin{table}[ht!]
\centering
{\small \setlength\tabcolsep{4.3pt} \begin{tabular}{|l|r|r|r|r|r|r|} \hline
& \multicolumn{3}{c|}{FR} & \multicolumn{3}{c|}{ES}  \\\cline{2-7} 
& \textbf{\em{H}}$\uparrow$ & \textbf{\em{D}}$\downarrow$ & Single & \textbf{\em{H}}$\uparrow$ & \textbf{\em{D}}$\downarrow$ & Single \\\hline
ORIG & \bf{75.20} & \bf{56.42} & \underline{79k} & \bf{78.42} & \bf{54.96} & \underline{92k} \\\hline
PB-SMT & 69.00 & 59.64 & 51k & 71.79 & 58.56 & 54k \\\hline
LSTM & 69.28 & 59.48 & 53k & 72.84 & 58.29 & 55k \\\hline
TRANS & 73.13 & 57.70 & 58k & 77.23 & 56.26 & 64k \\\hline\hline

& \multicolumn{3}{c|}{EN$_{FR}$} & \multicolumn{3}{c|}{EN$_{ES}$}\\\hline
& \textbf{\em{H}}$\uparrow$ & \textbf{\em{D}}$\downarrow$ & Single & \textbf{\em{H}}$\uparrow$ & \textbf{\em{D}}$\downarrow$ & Single \\\hline
ORIG & \textbf{59.04} & \textbf{63.43} & \underline{78k} & \textbf{59.05} & \bf{63.42} & \underline{78k}\\\hline
PB-SMT & 55.57 & 65.80 & 51k & 56.31 & 65.29 & 61k\\\hline
LSTM & 53.15 & 67.02 & 50k & 53.85 & 66.64 & 48k \\\hline
TRANS & 55.85 & 65.43 & 58k & 56.22 & 65.19 & 68k\\\hline
\end{tabular}
    \caption{Shannon entropy (\textbf{\em{H}}) for French, Spanish, English (EN$_{FR}$ and EN$_{ES}$) and Simpson's diversity index (\textbf{\em{D}}) for original training data and the output of the PB-SMT, LSTM and TRANS systems. Scores are multiplied by 100 for ease of readability.}
    \label{tbl:Shannonfrenchspanishenglish}}
\end{table}
The \textbf{\em{H}} and \textbf{\em{D}} scores in Table~\ref{tbl:Shannonfrenchspanishenglish} are an evidence of the negative impact of MT on the morphological diversity -- the scores for the ORIG indicate a consistent higher diversity. Comparing the MT systems, it results that TRANS retains morphological diversity better than the others. LSTM performs better than PB-SMT for translations into the morphologically richer languages (FR and ES) but PB-SMT seems much better than LSTM for translations into English. While the loss of lexical diversity could, in some cases be a desirable side-effect of MT systems (in terms of simplification or consistency), the uncontrolled loss of morphological richness is problematic as it can prevent systems from picking the grammatically correct option.

\section{Conclusions}
In this work, we explore the effects of MT algorithms on the richness and complexity of language. We establish that there is indeed a quantitatively measurable difference between the linguistic richness of MT systems' training data and their output -- a product of algorithmic bias. These findings are in line with previous results described in Vanmassenhove et al.~\shortcite{vanmassenhove-etal-2019-lost}. Assessing diversity or richness in language is a multifacted task spanning over various domains. As such, we approach this task from multiple angles focusing on lexical diversity and sophistication, morphological variety and a more translation specific metric focusing on synonymy. To do so, we analyse the results of 9 different metrics including established, newly proposed and adapted ones. The metrics suit we developed is unprecedented in the study of MT quality and we believe it could drive future research on MT evaluation. 

Based on a wide range of experiments with 3 different MT architectures, we draw the following main conclusions: (i) all 9 metrics indicate that the original training data has more lexical and morphological diversity compared to translations produced by the MT systems. This is the case for all language pairs and directions; (ii) Comparing the MT systems among themselves, there is a strong indication (for most metrics) that Transformer models outperform the others in terms of lexical and morphological richness. We also ought to note that, on average, the ranking of the systems in terms of diversity metrics correlates with the quality of the translations (in terms of BLEU and TER). This is something that would need to be further explored in future work; (iii) The data for the morphologically richer languages (ES, FR) has higher lexical and (evidently) morphological diversity than the English data both in the original data and in the translations generated by all systems. However, for PB-SMT, LSTM and TRANS the difference in scores is much smaller than the ORIG, indicating that the MT systems have a stronger negative impact (in terms of diversity and richness) on the morphologically richer languages.

\section{Acknowledgements}
We would like to thank the reviewers for their insightful comments and feedback.

%\section*{Acknowledgements}

\bibliography{eacl2021}
\bibliographystyle{acl_natbib}

\end{document}